\documentclass{article}

\usepackage{microtype}
\usepackage{graphicx}
\usepackage{booktabs} 

\usepackage{hyperref}

\usepackage{multicol}
\usepackage{multirow}
\usepackage{bm}
\usepackage{graphicx}
\usepackage{xcolor}
\usepackage{amsmath}
\usepackage{amsthm}
\usepackage{subcaption}

\usepackage[accepted]{icml2025}

\usepackage{amsmath}
\usepackage{amssymb}
\usepackage{mathtools}
\usepackage{amsthm}

\usepackage[capitalize,noabbrev]{cleveref}

\theoremstyle{plain}
\newtheorem{theorem}{Theorem}[section]

\theoremstyle{definition}
\newtheorem{definition}[theorem]{Definition}
\newtheorem{assumption}[theorem]{Assumption}
\theoremstyle{remark}
\newtheorem{remark}[theorem]{Remark}

\usepackage[textsize=tiny]{todonotes}

\icmltitlerunning{\textit{Rainy}: Unlocking Satellite Calibration for Deep Learning in Precipitation}

\begin{document}

\twocolumn[
\icmltitle{\textit{Rainy}: Unlocking Satellite Calibration for Deep Learning in Precipitation}



\icmlsetsymbol{corr}{*}

\begin{icmlauthorlist}
\icmlauthor{Zhenyu Yu}{yzy}
\icmlauthor{Hanqing Chen}{chq}
\icmlauthor{Mohd Yamani Idna Idris}{yzy}
\icmlauthor{Pei Wang}{corr,wp}
\end{icmlauthorlist}

\icmlaffiliation{wp}{Faculty of Information Engineering and Automation, Kunming University of Science and Technology, Kunming, 650500, China}
\icmlaffiliation{yzy}{Faculty of Computer Science and Information Technology, Universiti Malaya, Kuala Lumpur, 50603, Malaysia}
\icmlaffiliation{chq}{College of Ocean Engineering and Energy, Guangdong Ocean University, Zhanjiang, 524088, China}

\icmlcorrespondingauthor{Zhenyu Yu}{yuzhenyuyxl@foxmail.com}
\icmlcorrespondingauthor{Hanqing Chen}{hanqing@gdou.edu.cn}
\icmlcorrespondingauthor{Mohd Yamani Idna Idris}{yamani@um.edu.my}
\icmlcorrespondingauthor{Pei Wang}{peiwang0518@163.com}

\icmlkeywords{Deep Learning, Remote Sensing, Computer Vision, Satellite Calibration, Precipitation}

\vskip 0.3in
]




\begin{figure*}
    \centering
    \includegraphics[width=1\linewidth]{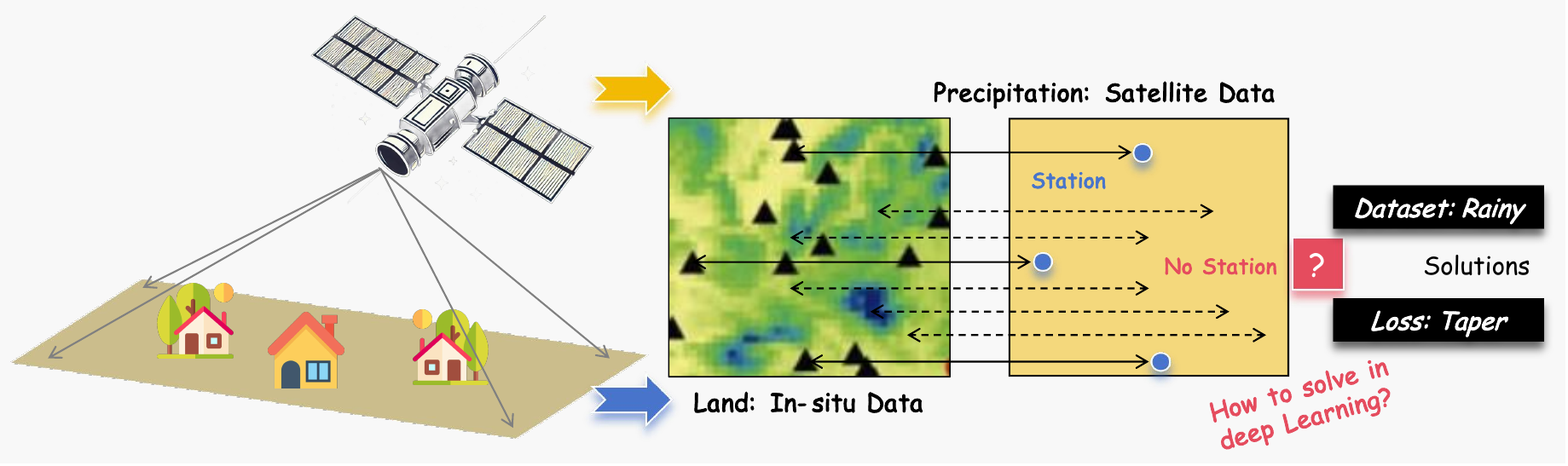}
    \caption{Motivation of this study. Satellite data provide wide coverage but lack ground-truth accuracy, while in-situ observations are sparse but reliable. The challenge is to integrate these data sources effectively for precipitation estimation. The proposed \textbf{\textit{Rainy}} dataset and \textbf{\textit{Taper Loss}} address this issue by leveraging deep learning to improve satellite calibration in regions with limited ground stations.}
    \label{fig_motivation}
\end{figure*}

\begin{abstract}
Precipitation plays a critical role in the Earth's hydrological cycle, directly affecting ecosystems, agriculture, and water resource management. Accurate precipitation estimation and prediction are crucial for understanding climate dynamics, disaster preparedness, and environmental monitoring. In recent years, artificial intelligence (AI) has gained increasing attention in quantitative remote sensing (QRS), enabling more advanced data analysis and improving precipitation estimation accuracy. Although traditional methods have been widely used for precipitation estimation, they face limitations due to the difficulty of data acquisition and the challenge of capturing complex feature relationships. Furthermore, the lack of standardized multi-source satellite datasets, and in most cases, the exclusive reliance on station data, significantly hinders the effective application of advanced AI models. To address these challenges, we propose the \textbf{\textit{Rainy}} dataset, a multi-source spatio-temporal dataset that integrates pure satellite data with station data, and propose \textbf{\textit{Taper Loss}}, designed to fill the gap in tasks where only in-situ data is available without area-wide support. The \textit{Rainy} dataset supports five main tasks: (1) satellite calibration, (2) precipitation event prediction, (3) precipitation level prediction, (4) spatiotemporal prediction, and (5) precipitation downscaling. For each task, we selected benchmark models and evaluation metrics to provide valuable references for researchers. Using precipitation as an example, the \textit{Rainy} dataset and \textit{Taper Loss} demonstrate the seamless collaboration between QRS and computer vision, offering data support for ``AI for Science'' in the field of QRS and providing valuable insights for interdisciplinary collaboration and integration.

\end{abstract}    
\section{Introduction}
Precipitation is one of the most critical components of the Earth's hydrological cycle, significantly impacting agriculture, ecosystems, and water resource management \cite{tsakiris2023adaptive, wang2023evidence}. Accurate and timely estimation of precipitation is essential for climate change studies, disaster management, and environmental monitoring \cite{tsatsaris2021geoinformation, boluwade2020remote, zhou2024enhancing}. Quantitative Remote Sensing (QRS) plays a vital role in estimating various physical parameters, including precipitation, through the analysis of satellite data \cite{abdalzaher2023early}. Traditionally, QRS methods rely on physical models and empirical algorithms, which, while effective in some cases, face challenges in handling complex, large-scale data due to their inherent assumptions and dependence on external parameters \cite{chen2022remote, yuan2020deep}.

Meanwhile, the field of Computer Vision (CV) has made rapid progress in recent years, fueled by deep learning algorithms and large datasets, leading to breakthroughs in image recognition, segmentation, and object detection tasks \cite{li2022deep}. However, despite significant advancements in CV, its application in quantitative remote sensing remains limited \cite{messina2022twenty}. This is mainly due to the lack of publicly available datasets specifically tailored for QRS tasks. Moreover, remote sensing imagery (e.g., satellite and hyperspectral data) differs significantly from RGB images commonly used in CV, making it challenging to directly apply existing CV models to QRS tasks.

\textbf{Motivation and existing problems.} One of the primary obstacles in integrating quantitative remote sensing with computer vision is the \textit{lack of standardized open datasets}, especially for quantitative tasks like precipitation estimation. In the QRS field, satellite data often require complex preprocessing steps, including atmospheric correction, geometric rectification, and cloud removal, to ensure data accuracy \cite{kucharczyk2021remote, tmuvsic2020current}. These preprocessing steps vary between studies, leading to inconsistencies in data quality and model performance, hindering fair comparisons across different methods \cite{zhao2022overview, schneider2010mapping}. Additionally, the issue of \textit{regional heterogeneity}—where environmental conditions vary significantly across geographic regions—means that models trained on data from one region may not generalize well to others \cite{mahrad2020contribution}. This lack of comparability between datasets further limits algorithm generalization and the development of robust models \cite{gorelick2017google}.

\textbf{Combining satellite and in-situ data.} In-situ data have the advantage of high accuracy and temporal resolution, providing precise local precipitation measurements but with limited spatial coverage \cite{kidd2020global}. In contrast, satellite data offer extensive spatial coverage with continuous observations over large areas, though they have relatively lower accuracy and resolution, often affected by atmospheric interference \cite{beck2017mswep}. Combining high-accuracy in-situ data with the broad coverage of satellite data can enable effective satellite calibration, achieving more accurate precipitation estimation \cite{le2023robustness, wu2020spatiotemporal}. Furthermore, deep learning algorithms have demonstrated strong performance in image recognition tasks, making them appealing for remote sensing applications like precipitation estimation. However, there is currently a lack of open datasets and tailored deep learning algorithms specifically designed to address the limitations posed by the scarcity of station data \cite{yuan2020deep, moraux2019deep}.

The main \textbf{contributions} of this work are as follows:

\begin{itemize}
    \item We present \textbf{\textit{Rainy}}, \textbf{a publicly available dataset} containing long-term, multi-source spatio-temporal data from both satellite and in-situ measurements. This dataset supports multiple tasks, including satellite calibration, event prediction, level prediction, spatio-temporal forecasting, and downscaling, providing benchmark data support for applying deep learning methods to satellite calibration in precipitation.

    \item We propose \textbf{a novel loss function}, \textbf{\textit{Taper Loss}}, designed to prioritize reliable in-situ data during the matching and calibration process between satellite and ground data, thereby significantly improving validation accuracy. By applying a distance-weighted mechanism, this objective function effectively suitable for tasks lacking area-wide data support.
    
    \item We conduct experiments with various machine learning models and deep learning models, establishing a standardized \textbf{benchmark} for integrating CV models with QRS tasks and ensuring fair comparisons across different models and methods. This dataset facilitates model training and evaluation, promoting the development of AI applications tailored to QRS.

\end{itemize}

\begin{figure*}[!ht]
    \centering
    \includegraphics[width=1\linewidth]{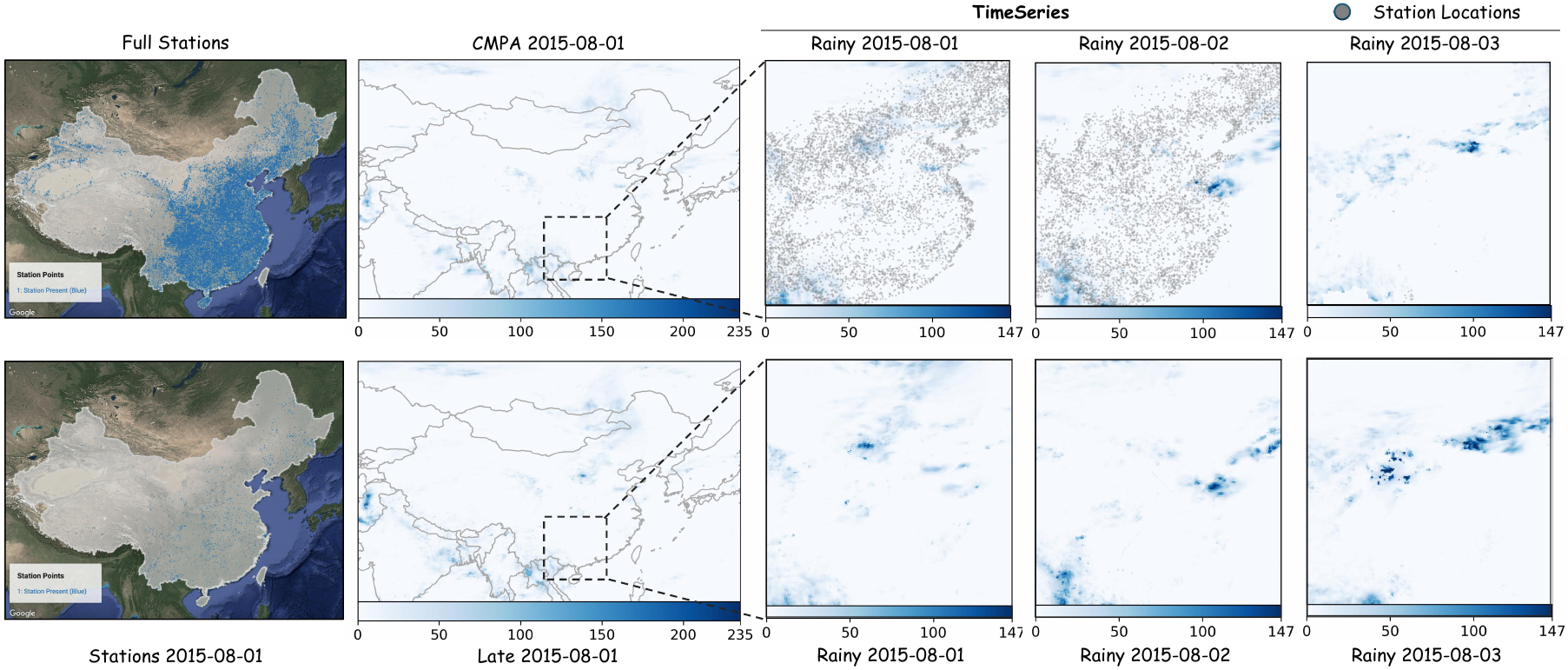}
    \caption{Overview of dataset \textit{Rainy}, including CMPA, CMPA station, and Late data. The \textbf{full data} represents the complete precipitation dataset over China (440 $\times$ 700 pixels), while the \textbf{partial data} corresponds to a selected region from the \textit{Rainy} dataset (256 $\times$ 256 pixels). This dataset is available in both daily and hourly versions, forming a long-term time series (0.1$^\circ$/pixel). The \textbf{full stations} dataset includes all stations across China from 2015 to 2017, though the specific stations corresponding to each data may vary. Notably, in CMPA data, only the stations provide accurate precipitation values. For certain time, CMPA data lacks station coverage, such as on August 3, 2015.}
    \label{fig_distribution}
\end{figure*}

\section{Related Work}

\subsection{Quantitative Remote Sensing}
Quantitative remote sensing involves analyzing spectral, spatial, and temporal information in remote sensing images to estimate physical and environmental parameters such as vegetation indices, soil moisture, and surface temperature \cite{wu2019advances}. Unlike qualitative analysis, quantitative remote sensing focuses on using mathematical models and physical equations to invert atmospheric and surface features, enabling precise monitoring and assessment of natural phenomena \cite{liang2005quantitative}. It has been widely applied in fields such as climate change monitoring, ecological evaluation, and agricultural management \cite{dalei2021advances}.

Satellite calibration is crucial in quantitative remote sensing to enhance the accuracy and reliability of satellite-derived measurements for applications such as climate monitoring, environmental analysis, and resource management \cite{xu2022onboard}. Calibration allows satellite data to closely align with ground-based observations by compensating for errors due to atmospheric conditions, sensor drift, and other environmental factors \cite{ma2021evaluation}. However, satellite calibration faces several challenges, including the limited spatial and temporal resolution of data and the varying atmospheric conditions that affect data quality. These issues underscore the need for robust calibration methods in quantitative remote sensing.

\subsection{Traditional Methods}
Traditional quantitative remote sensing methods primarily rely on physical and empirical models. Radiative Transfer Models (RTMs) are typical examples of physical models used to simulate the propagation of electromagnetic radiation through the atmosphere and surface, enabling the retrieval of surface characteristics \cite{himeur2022using, cheng2020remote}. Empirical models, such as Vegetation Indices (VIs), estimate surface vegetation cover based on the ratio of near-infrared to red reflectance \cite{spadoni2020analysis, liu2020similarity}.

Despite their utility in specific scenarios, these methods face several challenges. \textbf{Parameter dependence}: Physical models require precise input parameters, such as atmospheric composition, topography, and vegetation distribution. The uncertainty in these parameters can significantly impact retrieval accuracy, and they are often difficult to measure or acquire \cite{pu2000hyperspectral, singh2020hyperspectral}. \textbf{Limited generalization}: Many empirical models heavily depend on monitoring data, performing well in calibrated regions but often struggling to generalize to other geographical areas with different environmental conditions. \textbf{High cost of acquiring monitoring data}: Establishing monitoring stations requires significant human and material resources, and it cannot compensate for the lack of historical observation data. \textbf{Complex data processing}: Traditional methods demand extensive data processing, especially for high-dimensional multispectral or hyperspectral data, leading to increased computational costs \cite{liu2020similarity}.

\subsection{Machine Learning Methods}
In recent years, data-driven methods have gradually supplanted traditional models, with significant progress made in the application of machine learning (ML) and deep learning (DL) techniques in quantitative remote sensing. Models such as Support Vector Machines (SVM) \cite{cervantes2020comprehensive}, Random Forests (RF) \cite{antoniadis2021random}, and k-Nearest Neighbors (k-NN) \cite{bansal2022comparative} have been applied to handle high-dimensional remote sensing data and capture complex nonlinear relationships \cite{diaz2022machine, prodhan2022review}. However, these ML models rely heavily on feature engineering and face challenges in scalability and computational efficiency as remote sensing datasets continue to grow. Additionally, while deep learning methods (e.g., LSTM \cite{hochreiter1997long}, ConvLSTM \cite{sainath2015convolutional}, UNet \cite{ronneberger2015u}) can automatically learn features, they require large amounts of labeled data and often struggle with generalization across different regions and sensor types \cite{ma2019deep, li2022deep}.

\begin{table*}[!ht]
    \centering
    \caption{Overview of the IMERG-Late and CMPA data products used in the \textit{Rainy} dataset (2015-2017).}
    \resizebox{1.0\linewidth}{!}{
    \begin{tabular}{ccccc}
    \toprule
    \textbf{Data Product} & \textbf{Source} & \textbf{Spatial Resolution} & \textbf{Temporal Resolution} & \textbf{Coverage Area} \\ \midrule
    \textbf{IMERG-Late}   & GPM (Global Precipitation Measurement) & 0.1° & 30 minutes & Global \\ 
    \textbf{CMPA}         & China Meteorological Administration & 0.05° & 1 hour     & China \\ \bottomrule
    \end{tabular}
    }
    \label{table_data_info}
\end{table*}

\subsection{Integrating Computer Vision with Quantitative Remote Sensing}
Integrating computer vision (CV) with QRS presents both opportunities and challenges. The primary challenge is the \textbf{data problem}. Unlike CV, which benefits from standardized datasets (e.g., ImageNet \cite{deng2009imagenet}), quantitative remote sensing lacks open datasets suitable for DL training and evaluation. Satellite data require extensive preprocessing (e.g., atmospheric correction, geometric rectification, and cloud removal \cite{abdelbaki2022review}), leading to inconsistencies across studies. Moreover, satellite data are highly \textbf{region-dependent}, making generalization across different geographic regions difficult.

Another challenge is \textbf{algorithm development}. Remote sensing data contain complex spatio-temporal information and multiple spectral channels, limiting the direct applicability of conventional CV models. Significant adaptation is required to transform in-situ and satellite data into meaningful supervisory signals for ML-based satellite calibration.

To address these challenges, a publicly available dataset with standardized preprocessing is essential for integrating CV with QRS. A well-structured dataset would enable fair benchmarking, foster robust algorithm development, and advance the field \cite{salcedo2020machine}.

\begin{table*}[!ht]
    \centering
    \caption{Statistics for dataset \textit{Rainy}. In (b), values displayed as 0.00 are rounded, but the actual values are near zero, not exactly zero. We selected 0.00 as the minimum and 6.22 (hourly) / 38.48 (daily) as the maximum (\textbf{bold}) for normalization. The normalization formula is \(x' = \frac{x - x_{\text{min}}}{x_{\text{max}} - x_{\text{min}}}\), where \(x\), \(x_{\text{min}}\), \(x_{\text{max}}\), and \(x'\) represent the original value, minimum, maximum, and normalized value, respectively.} 
    \resizebox{0.9\textwidth}{!}{
    \begin{subtable}[t]{1.0\linewidth}
        \caption{Unfiltered data with all values included.}
        \centering
        \begin{tabular}{c|ccccccc|c|ccccccc}
        \toprule
            \textbf{Hourly} & \textbf{Min} & \textbf{Max} & \textbf{Avg} & \textbf{Q1} & \textbf{Q2} & \textbf{Q3}  & \textbf{Q99} & \textbf{Daily} & \textbf{Min} & \textbf{Max} & \textbf{Avg} & \textbf{Q1} & \textbf{Q2} & \textbf{Q3} & \textbf{Q99} \\ \midrule
            CMPA & 0.00  & 322.60  & 0.09  & 0.00  & 0.00  & 0.00  & 1.52  & CMPA & 0.00  & 761.15  & 1.90  & 0.00  & 0.00  & 0.11  & 23.25  \\ 
            Late & 0.00  & 120.00  & 0.12  & 0.00  & 0.00  & 0.00  & 2.11  & Late & 0.00  & 863.20  & 2.88  & 0.00  & 0.03  & 1.07  & 32.31  \\ 
            All & 0.00  & 322.60  & 0.11  & 0.00  & 0.00  & 0.00  & 1.81  & All & 0.00  & 863.20  & 2.39  & 0.00  & 0.00  & 0.44  & 27.27  \\ \bottomrule
        \end{tabular}
    \end{subtable}
    }
    \resizebox{0.9\textwidth}{!}{
    \begin{subtable}[t]{1.0\linewidth}
        \centering
        \caption{Data after removing zero values.}
        \begin{tabular}{c|ccccccc|c|ccccccc}
        \toprule
            \textbf{Hourly} & \textbf{Min} & \textbf{Max} & \textbf{Avg} & \textbf{Q1} & \textbf{Q2} & \textbf{Q3} & \textbf{Q99} & \textbf{Daily} & \textbf{Min} & \textbf{Max} & \textbf{Avg} & \textbf{Q1} & \textbf{Q2} & \textbf{Q3} & \textbf{Q99} \\ \midrule
            CMPA & 0.02  & 322.60  & 1.22  & 0.20  & 0.49  & 1.12  & 6.80  & CMPA & 0.02  & 761.15  & 6.92  & 0.49  & 1.98  & 6.58  & 37.25  \\ 
            Late & 0.00  & 120.00  & 0.75  & 0.02  & 0.12  & 0.52  & 5.80  & Late & 0.00  & 863.20  & 5.07  & 0.08  & 0.63  & 3.43  & 40.29  \\
            All & 0.00  & 322.60  & 0.90  & 0.08  & 0.26  & 0.77  & \textbf{6.22}  & All & 0.00  & 863.20  & 5.67  & 0.22  & 1.15  & 4.84  & \textbf{38.48}  \\ \bottomrule
        \end{tabular}
    \end{subtable}
    }
    \label{table_statis}
\end{table*}

\section{Proposed Dataset - \textit{Rainy}}



\subsection{Data Information}


\textbf{IMERG-Late} (Integrated Multi-satellite Retrievals for GPM - Late Run) is a satellite-based precipitation product, merging microwave, infrared, and radar data to generate high spatiotemporal global precipitation estimates. IMERG-Late data are particularly useful for short-term precipitation forecasting and real-time event prediction.\\
\textbf{CMPA} (China Meteorological Administration Precipitation Analysis) is a high-resolution precipitation product that integrates ground-based and satellite data, developed specifically for China. CMPA enhances the precision of regional precipitation estimates and helps calibrate satellite-based precipitation data, ensuring higher accuracy in localized analyses.\\
These datasets provide essential references for precipitation estimation, supporting both regional and global-scale analysis. Data information can be found in Table \ref{table_data_info}. Temporal and spatial distribution is shown in Figure \ref{fig_distribution}.

\subsection{Supported Tasks}

\textbf{Task 1: Satellite calibration.} 
Satellite calibration for precipitation involves estimating actual precipitation in a region using satellite data. Due to differences in sensor measurements, satellite data must be calibrated against ground station data. This task is framed as a regression problem, where the model learns the relationship between satellite data and ground observations to improve precipitation estimation accuracy. This is the \textbf{primary task}, which involves calibrating satellite data using \textit{Taper Loss} and station data. \\
    \textbf{Task Type}: Regression \\
    \textbf{Objective}: Precipitation amount (unit: mm) \\
    \textbf{Benchmark Models}: RFR \cite{segal2004machine}, BPANN \cite{hecht1992theory}, XGBoost \cite{chen2015xgboost}, MLP \cite{gardner1998artificial}, LSTM \cite{hochreiter1997long}, UNet \cite{ronneberger2015u}, and ConvLSTM \cite{sainath2015convolutional}.

\textbf{Task 2: Precipitation event prediction.}
The goal of precipitation event prediction is to determine whether it will rain in a specific region. Due to the nature of satellite data, where each pixel represents a geographical area, this task is a binary classification problem that predicts whether precipitation will occur in the region. The task requires accurately identifying precipitation events across a wide spatial area.\\
    \textbf{Task Type}: Binary classification \\
    \textbf{Objective}: Will it rain (0-No / 1-Yes) \\
    \textbf{Benchmark Models}: RF \cite{breiman2001random}, MLP \cite{gardner1998artificial}, BPANN \cite{hecht1992theory}, and LSTM \cite{hochreiter1997long}.

\textbf{Task 3: Precipitation level prediction.} 
The goal of precipitation level prediction is to classify precipitation into different levels based on the amount of rainfall. The precipitation level standards are detailed in Appendix. This is a multi-class classification problem, where the model predicts the precipitation intensity for a given region based on input satellite data.\\
    \textbf{Task Type}: Multi-class classification \\
    \textbf{Objective}: Precipitation level (0-No rain / 1-Light rain / 2-Moderate rain / 3-Heavy rain / 4-Storm / 5-Severe storm / 6-Extraordinary storm). The level standards are shown in Table.\\
    \textbf{Benchmark Models}: RF \cite{breiman2001random}, MLP \cite{gardner1998artificial}, BPANN \cite{hecht1992theory}, and LSTM \cite{hochreiter1997long}.

\textbf{Task 4: Spatiotemporal prediction.} 
Spatiotemporal prediction of precipitation aims to forecast the distribution of precipitation over time. This task requires capturing both spatial and temporal dynamics, making it a spatiotemporal prediction problem.\\
    \textbf{Task Type}: Spatio-temporal prediction \\
    \textbf{Objective}: Predict the future spatiotemporal distribution of precipitation across regions, incorporating both spatial patterns and temporal changes (unit: mm). \\
    \textbf{Benchmark Model}: ConvLSTM \cite{sainath2015convolutional}.

\textbf{Task 5: Precipitation downscaling.} 
Precipitation downscaling refers to the task of enhancing low-resolution satellite precipitation data to a higher resolution. This super-resolution reconstruction task typically relies on deep learning models to enhance spatial details and improve the resolution of precipitation data.\\
    \textbf{Task Type}: Super-resolution reconstruction \\
    \textbf{Objective}: Generate high-resolution precipitation data from low-resolution satellite observations to improve spatial accuracy and detail  (unit: mm). \\
    \textbf{Benchmark Models}: Bicubic \cite{parker2010algorithms}, SRCNN \cite{dong2015image}, ESPCN \cite{shi2016real}, LapSRN \cite{lai2017deep}, EDSR \cite{lim2017enhanced}, ESRGAN \cite{wang2018esrgan}, BSRGAN \cite{zhang2021designing}, SwinIR \cite{liang2021swinir},  and DiffIR \cite{xia2023diffir}.

\section{Method - \textit{Taper Loss}}


We propose \textit{Taper Loss}, a novel loss function that incorporates distance-based weighting via kernel functions to minimize the error between sparse ground observation points and satellite image predictions. This loss function is designed to handle remote sensing tasks where ground observations are sparse and unevenly distributed, thereby improving model prediction accuracy.

\begin{definition}[Satellite and ground data]
    Let $\mathbf{I}_1 \in \mathbb{R}^{M \times M}$ be the satellite data for the region of interest, and $\mathbf{I}_2 \in \mathbb{R}^{M \times M}$ be the ground observations, where only $N$ points are reliable while the rest are interpolated with uncertainty.
\end{definition}

\begin{definition}[Reliable observation stations]
    Let $\{(x_1, y_1), (x_2, y_2), \dots, (x_N, y_N)\}$ represent the coordinates of $N$ reliable ground observation points, with corresponding true values $\{z_1, z_2, \dots, z_N\}$.
\end{definition}

\begin{assumption}[Sparse observations]
    Due to the sparse distribution of observation points, directly generating a continuous spatial distribution map leads to inaccuracies. To address this, we propose \textit{Taper Loss}, a loss function that incorporates distance decay to focus on reliable data points and weights the loss based on proximity.
\end{assumption}

\begin{definition}[Distance-based kernel function]
    Let \( d_j \) be the distance from the $j$-th reliable ground observation point to its nearest observation, defined as:
    \begin{equation}
        d_j = d(x_j, y_j)
    \end{equation}
    The distance-based kernel function \( K(d_j) \) weights each point’s error according to its distance. Common kernel functions include:
    \begin{equation}
        K(d_j) = 
        \begin{cases}
        e^{-\alpha d_j}, & \text{(Exponential decay kernel)} \\
        \max(0, 1 - \beta d_j), & \text{(Linear decay kernel)} \\
        \frac{1}{d_j^\gamma}, & \text{(Power-law decay kernel)} \\
        e^{-\frac{d_j^2}{2\sigma^2}}, & \text{(Gaussian decay kernel)}
        \end{cases}
    \end{equation}
    The choice of kernel function can be tailored based on the specific requirements of the task to capture different types of distance decay behavior.
\end{definition}

\begin{theorem}[Taper loss function]
    The \textit{Taper Loss} is defined as a weighted mean squared error, where the kernel function \( K(d_j) \) down-weights the influence of interpolated points, focusing on minimizing the error at reliable points:
    \begin{equation}
        \mathcal{L}_{\text{Taper}} = 
        \sum_{j=1}^{N} K(d_j) 
        \cdot (\mathbf{I}_1(x_j, y_j) - z_j)^2.
    \end{equation}
    where \( K(d_j) \) is the kernel function weighting the error based on the distance \( d_j \) to the nearest reliable observation. \( (x_j, y_j) \) represents the coordinates of the \( j \)-th reliable ground observation point, with \( \mathbf{I}_1(x_j, y_j) \) as the predicted value from the satellite image and \( z_j \) as the corresponding true observation value.
\end{theorem}

\begin{remark}[Normalization for multiple reliable points]
    To ensure balanced contributions from multiple reliable observation points, we introduce a normalization factor to prevent any single point from dominating the loss calculation. The normalized \textit{Taper Loss} is defined as:
    \begin{equation}
        \mathcal{L}_{\text{Taper, General}} = 
        \sum_{j=1}^{N} \frac{K(d_j)}{\sum_{k=1}^{N} K(d_k)} 
        \cdot (\mathbf{I}_1(x_j, y_j) - z_j)^2.
    \end{equation}
    where \( \sum_{k=1}^{N} K(d_k) \) is the normalization factor, ensuring balanced influence across observation points.
\end{remark}

\begin{theorem}[Optimization objective]
    To minimize \textit{Taper Loss}, we define the optimization objective with respect to model parameters \( \theta \) as:
    \begin{equation}
        \min_{\theta} \mathcal{L}_{\text{Taper}}(\theta) = 
        \sum_{j=1}^{N} \frac{K(d_j)}{\sum_{k=1}^{N} K(d_k)} 
        \cdot (\mathbf{I}_1(x_j, y_j; \theta) - z_j)^2.
    \end{equation}
    By applying gradient descent or other optimization methods, the model can reduce errors in regions with reliable data while down-weighting the influence of less reliable areas.
\end{theorem}

\begin{definition}[Total loss function]
    The total loss function combines \textit{Taper Loss} with additional loss terms, formulated as:
    \begin{equation}        
        \mathcal{L}_{\text{Total}} = \alpha \mathcal{L}_{\text{Taper}} + \beta \mathcal{L}_{\text{Other}},
    \end{equation}
    where \( \mathcal{L}_{\text{Taper}} \) incorporates distance-based weighting to prioritize reliable observation points, and \( \mathcal{L}_{\text{Other}} \) represents commonly used loss functions. In this work, we consider \( \mathcal{L}_1 \) (Mean Absolute Error) and \( \mathcal{L}_2 \) (Mean Squared Error) as examples. The hyperparameters \( \alpha \) and \( \beta \) control the relative contributions of the two terms.     
    The hyperparameters \( \alpha \) and \( \beta \) balance \textit{Taper Loss} and other loss terms, with default values set to 1.
\end{definition}

\section{Benchmark}

\subsection{Experimental Settings}

The experiments were conducted on a single NVIDIA GeForce RTX A100 GPU with 80 GB of memory. For each task and model, hyperparameters were tuned to optimize performance. 
All machine learning models are initialized with default parameters, except for UNet and ConvLSTM in \textbf{Tasks 1 to 4}, where the learning rate is set to 0.001, the batch size to 32, and the number of epochs to 100. For \textbf{Task 5}, we downsample the data to 2$\times$ as low-resolution images, while the original data is used as high-resolution images. The learning rate is set to 1e-4, the batch size to 16, and the number of epochs to 200. To ensure the reliability of the experimental results, each experiment was run five times with different random seeds. If the standard deviations were all less than 0.05, then omitted for simplicity.

Due to the large range of precipitation values and the presence of many zero values, there is significant data imbalance (see Table \ref{table_statis}). To address this, we opted not to use z-score or standard min-max normalization methods. Instead, we chose a 99$^{th}$ percentile (Q99) truncation normalization approach, which helps prevent low values from being overly diminished. The data for Task 5 is normalized to a range of 0-255 using the corresponding data's minimum and maximum values. This task does not require consideration of the issue of excessive zero values.


\subsection{Evaluation Metrics}

For each task in the \textit{Rainy} dataset, specific evaluation metrics are applied. \textbf{Task 1} is evaluated using mean squared error (MSE) \cite{chai2014root}, mean absolute error (MAE) \cite{willmott2005advantages}, and the coefficient of determination ($R^2$) \cite{willmott2012refined}. For \textbf{Task 2} and \textbf{Task 3}, standard classification metrics including Accuracy \cite{chicco2020advantages}, Precision \cite{nasir2021deep}, Recall \cite{kleyko2018classification}, and F1-score \cite{chicco2020advantages} are used. \textbf{Task 4} employs RMSE \cite{chai2014root} and MAE to measure predictive accuracy. For \textbf{Task 5}, we adopt RMSE, MAE, peak signal-to-noise ratio (PSNR) \cite{hore2010image}, and structural similarity index (SSIM) \cite{wang2004image} to assess both numerical accuracy and reconstruction quality. It should be noted that for Task 1, we have other metrics to choose from, which are commonly used in the field of satellite sensor calibration, as shown in the Appendix.

\begin{table*}[!ht]
    \centering
    \caption{Comparison of different methods for Task 2 (Precipitation Event Prediction) and Task 3 (Precipitation Level Prediction). \textbf{Bold} values indicate the best performance, while \underline{underlined} values represent the second-best results.}
    \resizebox{0.85\linewidth}{!}{
    \begin{tabular}{c|cccc|cccc}
    \toprule
        \multirow{2}{*}{\textbf{Method}} & \multicolumn{4}{c|}{\textbf{Task 2}} & \multicolumn{4}{c}{\textbf{Task 3}} \\ \cline{2-9}
        ~ & \textbf{Accuracy} & \textbf{Precision} & \textbf{Recall} & \textbf{F1 Score} & \textbf{Accuracy} & \textbf{Precision} & \textbf{Recall} & \textbf{F1 Score}  \\ \midrule
        RF & 0.6936 & 0.6751 & 0.6936 & 0.6705 & 0.6405 & 0.5877 & 0.6405 & 0.6036  \\ 
        MLP & \textbf{0.7181} & \underline{0.707} & \textbf{0.7181} & \underline{0.7079} & \textbf{0.6746} & \underline{0.5936} & \textbf{0.6746} & \underline{0.6223}  \\ 
        BPANN & \textbf{0.7181} & 0.7067 & \textbf{0.7181} & 0.7072 & \underline{0.6745} & 0.5918 & \underline{0.6745} & 0.6212  \\ 
        LSTM & \underline{0.7178} & \textbf{0.7075} & \underline{0.7178} & \textbf{0.7088} & 0.6739 & \textbf{0.5997} & 0.6739 & \textbf{0.6316} \\ 
        \bottomrule
    \end{tabular}
    }
    \label{table_compare_task2_3}
\end{table*}

\subsection{Comparison in Different Tasks}

For \textbf{Task 1 (Satellite calibration)}, as shown in Table \ref{table_compare_task1}, we evaluated models on both daily and hourly precipitation data. The results indicate that the UNet model combined with \textit{Taper Loss} achieved the best performance, with significant improvements in MSE, MAE, and $R^2$ compared to other models. Specifically, UNet$+\mathcal{L}_{Taper}$ outperformed other methods with an MSE of 0.1349 and 0.1174, MAE of 0.045 and 0.0179, and $R^2$ of 0.6339 and 0.5311 for daily and hourly data, respectively. The results demonstrate the effectiveness of using \textit{Taper Loss} for satellite calibration tasks, particularly in improving model accuracy and robustness.

For \textbf{Task 2 and Task 3 (Precipitation event and level prediction)} (see Table \ref{table_compare_task2_3}), MLP achieved the highest accuracy and F1-score for both tasks, with slight variations between metrics. Specifically, MLP and BPANN performed competitively, with accuracies around 0.7181 for Task 2 and 0.6746 for Task 3. LSTM also exhibited strong performance, particularly in Precision and F1-score. These results suggest that deep learning models such as MLP and LSTM are well-suited for precipitation classification tasks, offering robust generalization across different scenarios.

In \textbf{Task 4 (Spatiotemporal prediction)}, where spatiotemporal prediction was evaluated (see Table \ref{table_compare_task4}). Models using \textit{Taper Loss} consistently outperformed those using $\mathcal{L}_1$ or $\mathcal{L}_2$ loss functions alone. The best results were obtained with $\mathcal{L}_{Taper} + \mathcal{L}_1$, achieving RMSE values of 0.2118 and 0.0958 and MAE values of 0.0666 and 0.0135 for daily and hourly data, respectively. These findings highlight the advantage of incorporating \textit{Taper Loss} in spatiotemporal prediction, particularly in reducing errors and enhancing temporal consistency.

For \textbf{Task 5 (Downscaling)}, we conducted experiments on 2$\times$ downscaling (see Table \ref{table_compare_task5}). Among the evaluated models, DiffIR achieved the best performance, with an RMSE of 0.0533, MAE of 0.0061, PSNR of 35.5604, and SSIM of 0.9855. SwinIR also demonstrated competitive performance, closely following DiffIR in all metrics. Traditional methods such as Bicubic interpolation showed significantly lower performance compared to deep learning-based models like SwinIR and DiffIR. These results indicate that advanced deep learning models are highly effective for precipitation downscaling tasks, offering superior accuracy and image reconstruction quality.

\textbf{Key Insights.}
The introduction of \textit{Taper Loss} significantly enhances model performance in tasks involving satellite calibration and spatiotemporal prediction, demonstrating its robustness in scenarios with only in-situ data. Deep learning models such as UNet, LSTM, and MLP consistently outperform traditional methods in both regression and classification tasks, highlighting the importance of leveraging advanced models for precipitation estimation. For downscaling tasks, state-of-the-art models like DiffIR and SwinIR exhibit remarkable improvements in accuracy and reconstruction quality, underscoring the potential of deep learning in high-resolution precipitation prediction. Overall, the experimental results validate the effectiveness of the \textit{Rainy} dataset and the proposed methods, providing a comprehensive benchmark for future research in quantitative remote sensing.

\begin{table}[!ht]
    \centering
    \caption{Comparison of different methods for Task 1 (Satellite Calibration) on daily and hourly data. \textbf{Bold} indicates the best performance, while \underline{underlined} represents the second-best results.} 
    \resizebox{1.0\linewidth}{!}{
    \begin{tabular}{c|ccc|ccc}
    \toprule
        \multirow{2}{*}{\textbf{Method}} & \multicolumn{3}{c|}{\textbf{Daily}} & \multicolumn{3}{c}{\textbf{Hourly}} \\ \cline{2-7}
        ~ & \textbf{MSE} & \textbf{MAE} & \textbf{R$^2$} & \textbf{MSE} & \textbf{MAE} & \textbf{R$^2$} \\ \midrule
        BPANN & 0.1988 & 0.0497 & 0.4887 & 0.1788 & 0.0443 & 0.4367 \\ 
        LSTM & 0.1982 & 0.0523 & 0.4899 & 0.1784 & 0.0463 & 0.4377 \\ 
        MLP & 0.1986 & 0.0510 & 0.4892 & 0.1788 & 0.0446 & 0.4368 \\ 
        RFR & 0.2952 & 0.0604 & 0.3897 & 0.2648 & 0.0529 & 0.3267 \\ 
        XGBoost & 0.2022 & 0.0514 & 0.4811 & 0.1816 & 0.0452 & 0.4302 \\ 
        ConvLSTM & 0.1713 & 0.0536 & 0.3883 & 0.1411 & 0.0472 & 0.3758 \\ 
        $+\mathcal{L}_{Taper}$ & 0.1538 & \underline{0.0455} & 0.5067 & 0.1317 & 0.0351 & \underline{0.4948} \\ 
        UNet & \underline{0.1429} & 0.0468 & \underline{0.5733} & \underline{0.1215} & \underline{0.0204} & 0.4736 \\ 
        $+\mathcal{L}_{Taper}$ & \textbf{0.1349} & \textbf{0.0450} & \textbf{0.6339} & \textbf{0.1174} & \textbf{0.0179} & \textbf{0.5311} \\ \bottomrule
    \end{tabular}
    }
    \label{table_compare_task1}
\end{table}

\begin{figure*}[!ht]
    \centering
    \begin{minipage}{0.50\textwidth}
        \centering 
        \includegraphics[width=\textwidth]{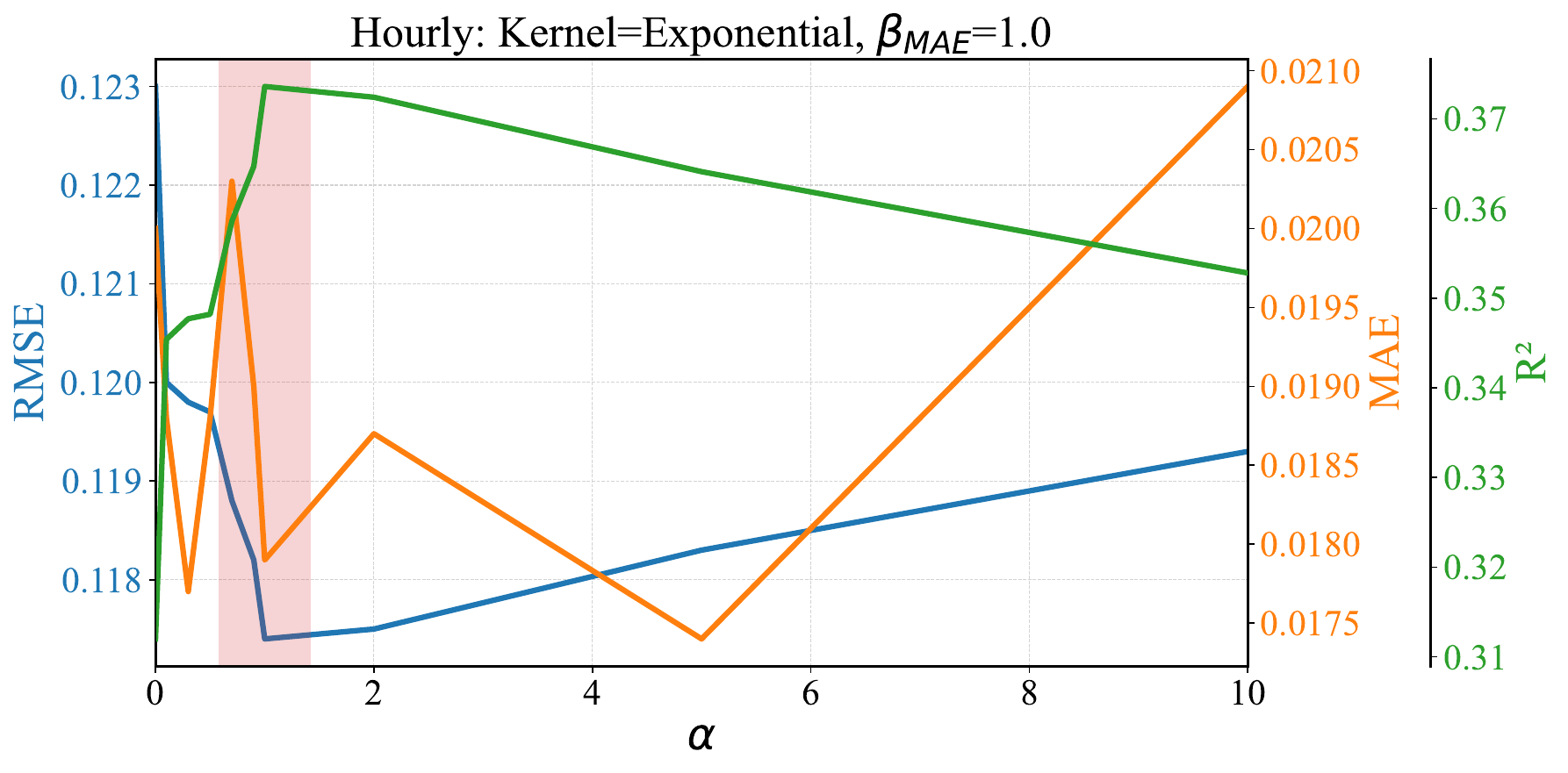}
    \end{minipage}
    \hspace{-6pt}
    \begin{minipage}{0.50\textwidth}
        \centering 
        \includegraphics[width=\textwidth]{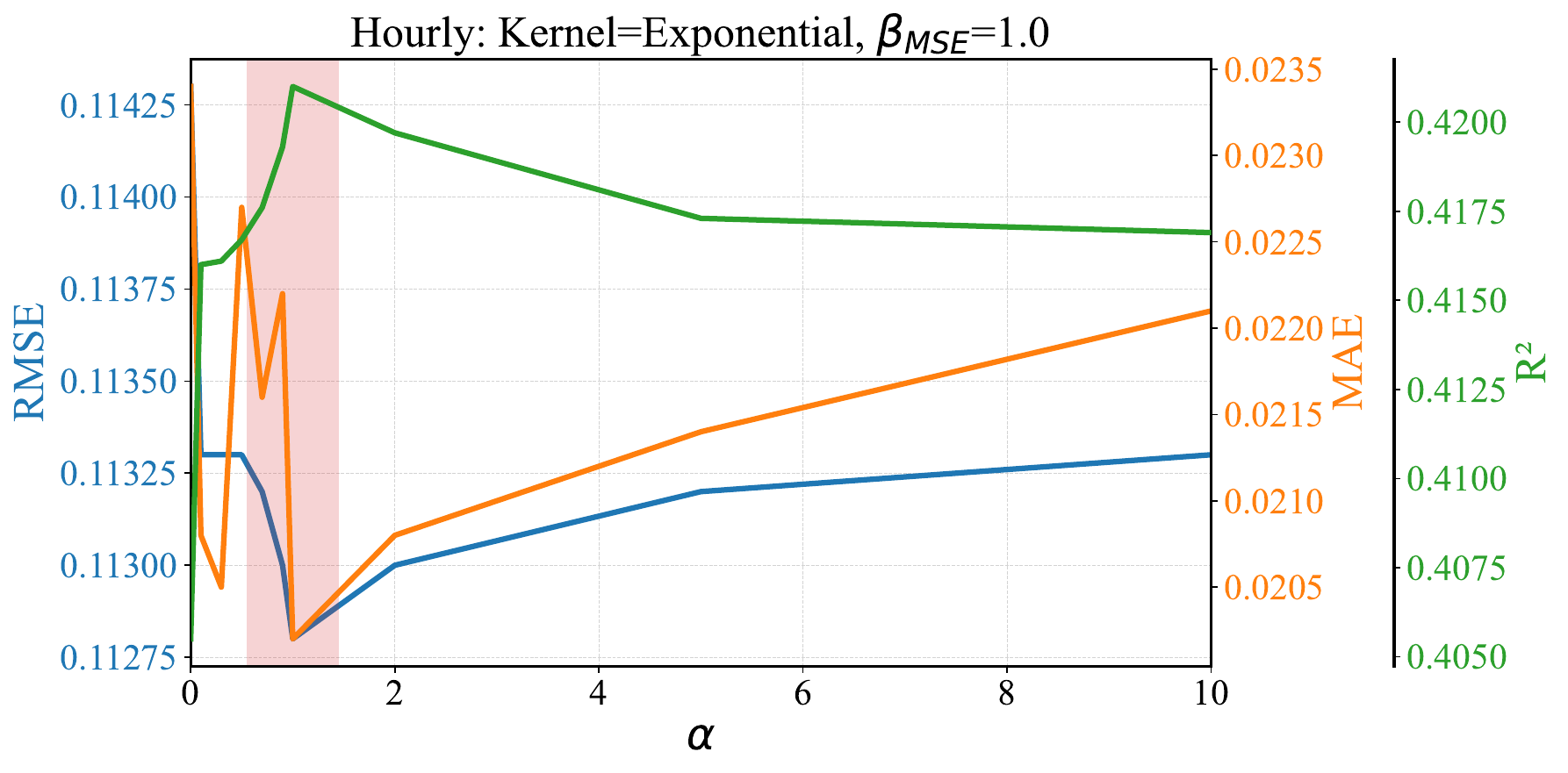}
    \end{minipage}
    \hspace{-6pt}
    \\
    \caption{Ablation study of the \textit{taper loss} parameter \( \alpha \) for the Hourly Data, using Exponential function as examples, evaluating the impact on RMSE, MAE, and \( R^2 \). The parameter \( \beta \) is set to 1. }
    \label{fig_alpha}
\end{figure*}

\begin{table}[!ht]
    \centering
    \caption{Comparison of different loss functions for Task 4 (Spatiotemporal Precipitation Prediction) on daily and hourly data. \textbf{Bold} values indicate the best performance, while \underline{underlined} values represent the second-best results.}
    \resizebox{0.95\linewidth}{!}{
    \begin{tabular}{c|cc|cc}
    \toprule
        \multirow{2}{*}{\textbf{Loss}} & 
        \multicolumn{2}{c|}{\textbf{Daily}} & \multicolumn{2}{c}{\textbf{Hourly}} \\ \cline{2-5}
        ~ &  \textbf{RMSE} &  \textbf{MAE} &  \textbf{RMSE} &   \textbf{MAE}  \\ \midrule
        $\mathcal{L}_{Taper} + \mathcal{L}_2$ & \underline{0.2255} & \underline{0.0754} & \underline{0.0987} & \underline{0.0162}  \\ 
        $\mathcal{L}_{Taper} + \mathcal{L}_1$ & \textbf{0.2118} & \textbf{0.0666} & \textbf{0.0958} & \textbf{0.0135} \\ \bottomrule
    \end{tabular}
    }
    \label{table_compare_task4}
\end{table}

\begin{table}[!ht]
    \centering
    \caption{Comparison of different methods for Task 5 (Precipitation Downscaling, \(2\times\)). \textbf{Bold} values indicate the best performance, while \underline{underlined} values represent the second-best results.}
    \resizebox{0.95\linewidth}{!}{
    \begin{tabular}{cccccc}
    \toprule
        \textbf{Method} & \textbf{RMSE} & \textbf{MAE} & \textbf{PSNR} & \textbf{SSIM} \\ \midrule
        Bicubic & 0.1621 & 0.0347 & 32.3905 & 0.9549  \\ 
        SRCNN & 0.1275 & 0.0227 & 34.2102 & 0.9554  \\ 
        ESPCN & 0.1174 & 0.0221 & 34.2533 & 0.9656  \\ 
        LapSRN & 0.1075 & 0.0187 & 35.2267 & 0.9654  \\ 
        EDSR & 0.0939 & 0.0134 & 34.3801 & 0.9775  \\ 
        ESRGAN & 0.0879 & 0.0139 & 34.4451 & 0.9718  \\ 
        BSRGAN & 0.0853 & 0.0114 & 35.0409 & 0.9745  \\ 
        SwinIR & \underline{0.0634} & \underline{0.0079} & \underline{35.5452} & \underline{0.9813}  \\ 
        DiffIR & \textbf{0.0533} & \textbf{0.0061} & \textbf{35.5604} & \textbf{0.9855} \\ \bottomrule
    \end{tabular}
    }
    \label{table_compare_task5}
\end{table}

\subsection{Ablation Study}
\textbf{Kernel function analysis.}  
Taking the satellite calibration task (Task 1) as an example, we analyze the impact of different kernel functions under two types of loss functions: \( \mathcal{L}_{Taper} + \mathcal{L}_1 \) and \( \mathcal{L}_{Taper} + \mathcal{L}_2 \). Table \ref{table_kernel_compare} presents the comparison results for daily and hourly data across four kernel types: Exponential, Linear, Power-law, and Gaussian.

\textbf{Daily data.}  
Under \( \mathcal{L}_{Taper} + \mathcal{L}_1 \), the Exponential kernel achieves the best RMSE of \( 0.1349 \pm 0.0049 \) and the highest \( R^2 \) of \( 0.6339 \pm 0.0232 \), demonstrating its superior capability in capturing spatial dependencies.  
For \( \mathcal{L}_{Taper} + \mathcal{L}_2 \), the Gaussian kernel performs best, with an RMSE of \( 0.133 \pm 0.0029 \) and an \( R^2 \) of \( 0.642 \pm 0.013 \), indicating its robustness in modeling complex spatial relationships.  
The Power-law kernel shows higher RMSE and MAE values, particularly under \( \mathcal{L}_{Taper} + \mathcal{L}_1 \), indicating lower consistency in performance and higher sensitivity to parameter variations.

\textbf{Hourly data.}  
For hourly data, under \( \mathcal{L}_{Taper} + \mathcal{L}_1 \), the Exponential kernel again achieves the best performance, with an RMSE of \( 0.1174 \pm 0.0016 \), an MAE of \( 0.0179 \pm 0.0012 \), and an \( R^2 \) of \( 0.3736 \pm 0.0179 \).  
Under \( \mathcal{L}_{Taper} + \mathcal{L}_2 \), the Exponential kernel continues to perform well, achieving the lowest RMSE of \( 0.1128 \pm 0.0004 \) and the highest \( R^2 \) of \( 0.421 \pm 0.0042 \). The Linear and Gaussian kernels show competitive performance, with slight variations in RMSE and MAE.  
Similar to daily data, the Power-law kernel exhibits the lowest consistency, confirming its sensitivity to parameter changes and its limited suitability for this task.

\textbf{Analysis of taper loss parameter \( \bm{\alpha} \).}  
Figure \ref{fig_alpha}, \ref{fig_alpha_daily}, and \ref{fig_alpha_hourly} illustrate the effect of the taper loss parameter \( \alpha \) on model performance for hourly data, using the Exponential kernel as an example. We evaluate the impact of \( \alpha \) on RMSE, MAE, and \( R^2 \), with the parameter \( \beta \) fixed at 1.

For both daily and hourly datasets, smaller values of \( \alpha \) (\( \alpha < 2 \)) lead to unstable performance, with significant fluctuations in RMSE, MAE, and \( R^2 \). This suggests that overly restrictive spatial tapering limits the model’s ability to generalize.  
At \( \alpha = 1 \), the model achieves the lowest RMSE and MAE, while \( R^2 \) approaches its peak, indicating that \( \alpha = 1 \) is an optimal hyperparameter setting that effectively balances bias and variance.  
As \( \alpha \) increases beyond 2, model performance stabilizes, showing that moderate tapering can maintain a good balance between bias and variance.  
Among all kernel functions, the Exponential and Gaussian kernels demonstrate smoother performance trends as \( \alpha \) increases, whereas the Power-law kernel remains more sensitive to changes in \( \alpha \).

In all experiments, the weight parameter \( \beta \) for the $\mathcal{L}_1$ and $\mathcal{L}_2$ is fixed at 1, ensuring that $\mathcal{L}_1$ or $\mathcal{L}_2$ plays a primary role in optimizing the objective function. This fixed value ensures consistent and reliable comparison across different kernel functions and loss types.


\section{Conclusion}




The main contributions of this work include the introduction of the \textit{Rainy} dataset and a novel loss function, \textit{Taper Loss}, designed to advance the integration of computer vision (CV) models with quantitative remote sensing (QRS) tasks. The \textit{Rainy} dataset integrates multi-source satellite data, including infrared, microwave, and radar observations, along with long-term ground station measurements, providing a comprehensive benchmark for spatio-temporal precipitation pattern analysis. By offering a standardized evaluation framework, \textit{Rainy} facilitates fair comparison, robust model training, and the development of AI-driven solutions for QRS. Additionally, the proposed \textit{Taper Loss} function effectively calibrates ground station data with spatial satellite observations, significantly improving the accuracy of loss calculation and model validation. This work represents a critical step in bridging the fields of CV and QRS, enabling more precise precipitation estimation and fostering innovative applications in environmental monitoring.


\section*{Impact Statement}
This study introduces \textbf{\textit{Rainy}}, a publicly available dataset based on deep learning, and proposes \textbf{\textit{Taper Loss}}, a novel approach to address the prediction challenges between station data and image data. The dataset is designed for five different spatiotemporal prediction tasks, including precipitation event prediction and precipitation estimation. The proposed dataset and method can be extended to a broader range of spatiotemporal forecasting applications, providing more accurate data support for environmental modeling and decision-making in weather-sensitive industries, thereby enhancing the reliability and practicality of meteorological predictions.


\bibliography{output}

\begin{thebibliography}{66}
\providecommand{\natexlab}[1]{#1}
\providecommand{\url}[1]{\texttt{#1}}
\expandafter\ifx\csname urlstyle\endcsname\relax
  \providecommand{\doi}[1]{doi: #1}\else
  \providecommand{\doi}{doi: \begingroup \urlstyle{rm}\Url}\fi

\bibitem[Abdalzaher et~al.(2023)Abdalzaher, Krichen, Yiltas-Kaplan, Ben~Dhaou, and Adoni]{abdalzaher2023early}
Abdalzaher, M.~S., Krichen, M., Yiltas-Kaplan, D., Ben~Dhaou, I., and Adoni, W. Y.~H.
\newblock Early detection of earthquakes using iot and cloud infrastructure: A survey.
\newblock \emph{Sustainability}, 15\penalty0 (15):\penalty0 11713, 2023.

\bibitem[Abdelbaki \& Udelhoven(2022)Abdelbaki and Udelhoven]{abdelbaki2022review}
Abdelbaki, A. and Udelhoven, T.
\newblock A review of hybrid approaches for quantitative assessment of crop traits using optical remote sensing: research trends and future directions.
\newblock \emph{Remote Sensing}, 14\penalty0 (15):\penalty0 3515, 2022.

\bibitem[Antoniadis et~al.(2021)Antoniadis, Lambert-Lacroix, and Poggi]{antoniadis2021random}
Antoniadis, A., Lambert-Lacroix, S., and Poggi, J.-M.
\newblock Random forests for global sensitivity analysis: A selective review.
\newblock \emph{Reliability Engineering \& System Safety}, 206:\penalty0 107312, 2021.

\bibitem[Bansal et~al.(2022)Bansal, Goyal, and Choudhary]{bansal2022comparative}
Bansal, M., Goyal, A., and Choudhary, A.
\newblock A comparative analysis of k-nearest neighbor, genetic, support vector machine, decision tree, and long short term memory algorithms in machine learning.
\newblock \emph{Decision Analytics Journal}, 3:\penalty0 100071, 2022.

\bibitem[Beck et~al.(2017)Beck, Van~Dijk, Levizzani, Schellekens, Miralles, Martens, and De~Roo]{beck2017mswep}
Beck, H.~E., Van~Dijk, A.~I., Levizzani, V., Schellekens, J., Miralles, D.~G., Martens, B., and De~Roo, A.
\newblock Mswep: 3-hourly 0.25 global gridded precipitation (1979--2015) by merging gauge, satellite, and reanalysis data.
\newblock \emph{Hydrology and Earth System Sciences}, 21\penalty0 (1):\penalty0 589--615, 2017.

\bibitem[Boluwade(2020)]{boluwade2020remote}
Boluwade, A.
\newblock Remote sensed-based rainfall estimations over the east and west africa regions for disaster risk management.
\newblock \emph{ISPRS journal of photogrammetry and remote sensing}, 167:\penalty0 305--320, 2020.

\bibitem[Breiman(2001)]{breiman2001random}
Breiman, L.
\newblock Random forests.
\newblock \emph{Machine learning}, 45:\penalty0 5--32, 2001.

\bibitem[Cervantes et~al.(2020)Cervantes, Garcia-Lamont, Rodr{\'\i}guez-Mazahua, and Lopez]{cervantes2020comprehensive}
Cervantes, J., Garcia-Lamont, F., Rodr{\'\i}guez-Mazahua, L., and Lopez, A.
\newblock A comprehensive survey on support vector machine classification: Applications, challenges and trends.
\newblock \emph{Neurocomputing}, 408:\penalty0 189--215, 2020.

\bibitem[Chai \& Draxler(2014)Chai and Draxler]{chai2014root}
Chai, T. and Draxler, R.~R.
\newblock Root mean square error (rmse) or mean absolute error (mae)?--arguments against avoiding rmse in the literature.
\newblock \emph{Geoscientific model development}, 7\penalty0 (3):\penalty0 1247--1250, 2014.

\bibitem[Chen et~al.(2022)Chen, Chen, Fu, Li, Jiang, Wang, Peng, Jia, and Hicks]{chen2022remote}
Chen, J., Chen, S., Fu, R., Li, D., Jiang, H., Wang, C., Peng, Y., Jia, K., and Hicks, B.~J.
\newblock Remote sensing big data for water environment monitoring: current status, challenges, and future prospects.
\newblock \emph{Earth's Future}, 10\penalty0 (2):\penalty0 e2021EF002289, 2022.

\bibitem[Chen(2015)]{chen2015xgboost}
Chen, T.
\newblock Xgboost: extreme gradient boosting.
\newblock \emph{R package version 0.4-2}, 1\penalty0 (4), 2015.

\bibitem[Cheng et~al.(2020)Cheng, Xie, Han, Guo, and Xia]{cheng2020remote}
Cheng, G., Xie, X., Han, J., Guo, L., and Xia, G.-S.
\newblock Remote sensing image scene classification meets deep learning: Challenges, methods, benchmarks, and opportunities.
\newblock \emph{IEEE Journal of Selected Topics in Applied Earth Observations and Remote Sensing}, 13:\penalty0 3735--3756, 2020.

\bibitem[Chicco \& Jurman(2020)Chicco and Jurman]{chicco2020advantages}
Chicco, D. and Jurman, G.
\newblock The advantages of the matthews correlation coefficient (mcc) over f1 score and accuracy in binary classification evaluation.
\newblock \emph{BMC genomics}, 21:\penalty0 1--13, 2020.

\bibitem[Dalei et~al.(2021)Dalei, Qing, Jianguang, Dongqin, Xiaodan, Xingwen, and Shengbiao]{dalei2021advances}
Dalei, H., Qing, X., Jianguang, W., Dongqin, Y., Xiaodan, W., Xingwen, L., and Shengbiao, W.
\newblock Advances in upscaling methods of quantitative remote sensing.
\newblock \emph{National Remote Sensing Bulletin}, 22\penalty0 (3):\penalty0 408--423, 2021.

\bibitem[Deng et~al.(2009)Deng, Dong, Socher, Li, Li, and Fei-Fei]{deng2009imagenet}
Deng, J., Dong, W., Socher, R., Li, L.-J., Li, K., and Fei-Fei, L.
\newblock Imagenet: A large-scale hierarchical image database.
\newblock In \emph{2009 IEEE conference on computer vision and pattern recognition}, pp.\  248--255. Ieee, 2009.

\bibitem[Diaz-Gonzalez et~al.(2022)Diaz-Gonzalez, Vuelvas, Correa, Vallejo, and Patino]{diaz2022machine}
Diaz-Gonzalez, F.~A., Vuelvas, J., Correa, C.~A., Vallejo, V.~E., and Patino, D.
\newblock Machine learning and remote sensing techniques applied to estimate soil indicators--review.
\newblock \emph{Ecological Indicators}, 135:\penalty0 108517, 2022.

\bibitem[Dong et~al.(2015)Dong, Loy, He, and Tang]{dong2015image}
Dong, C., Loy, C.~C., He, K., and Tang, X.
\newblock Image super-resolution using deep convolutional networks.
\newblock \emph{IEEE transactions on pattern analysis and machine intelligence}, 38\penalty0 (2):\penalty0 295--307, 2015.

\bibitem[Gardner \& Dorling(1998)Gardner and Dorling]{gardner1998artificial}
Gardner, M.~W. and Dorling, S.
\newblock Artificial neural networks (the multilayer perceptron)—a review of applications in the atmospheric sciences.
\newblock \emph{Atmospheric environment}, 32\penalty0 (14-15):\penalty0 2627--2636, 1998.

\bibitem[Gorelick et~al.(2017)Gorelick, Hancher, Dixon, Ilyushchenko, Thau, and Moore]{gorelick2017google}
Gorelick, N., Hancher, M., Dixon, M., Ilyushchenko, S., Thau, D., and Moore, R.
\newblock Google earth engine: Planetary-scale geospatial analysis for everyone.
\newblock \emph{Remote sensing of Environment}, 202:\penalty0 18--27, 2017.

\bibitem[Hecht-Nielsen(1992)]{hecht1992theory}
Hecht-Nielsen, R.
\newblock Theory of the backpropagation neural network.
\newblock In \emph{Neural networks for perception}, pp.\  65--93. Elsevier, 1992.

\bibitem[Himeur et~al.(2022)Himeur, Rimal, Tiwary, and Amira]{himeur2022using}
Himeur, Y., Rimal, B., Tiwary, A., and Amira, A.
\newblock Using artificial intelligence and data fusion for environmental monitoring: A review and future perspectives.
\newblock \emph{Information Fusion}, 86:\penalty0 44--75, 2022.

\bibitem[Hochreiter(1997)]{hochreiter1997long}
Hochreiter, S.
\newblock Long short-term memory.
\newblock \emph{Neural Computation MIT-Press}, 1997.

\bibitem[Hore \& Ziou(2010)Hore and Ziou]{hore2010image}
Hore, A. and Ziou, D.
\newblock Image quality metrics: Psnr vs. ssim.
\newblock In \emph{2010 20th international conference on pattern recognition}, pp.\  2366--2369. IEEE, 2010.

\bibitem[Kidd et~al.(2020)Kidd, Takayabu, Skofronick-Jackson, Huffman, Braun, Kubota, and Turk]{kidd2020global}
Kidd, C., Takayabu, Y.~N., Skofronick-Jackson, G.~M., Huffman, G.~J., Braun, S.~A., Kubota, T., and Turk, F.~J.
\newblock The global precipitation measurement (gpm) mission.
\newblock \emph{Satellite Precipitation Measurement: Volume 1}, pp.\  3--23, 2020.

\bibitem[Kleyko et~al.(2018)Kleyko, Rahimi, Rachkovskij, Osipov, and Rabaey]{kleyko2018classification}
Kleyko, D., Rahimi, A., Rachkovskij, D.~A., Osipov, E., and Rabaey, J.~M.
\newblock Classification and recall with binary hyperdimensional computing: Tradeoffs in choice of density and mapping characteristics.
\newblock \emph{IEEE transactions on neural networks and learning systems}, 29\penalty0 (12):\penalty0 5880--5898, 2018.

\bibitem[Kucharczyk \& Hugenholtz(2021)Kucharczyk and Hugenholtz]{kucharczyk2021remote}
Kucharczyk, M. and Hugenholtz, C.~H.
\newblock Remote sensing of natural hazard-related disasters with small drones: Global trends, biases, and research opportunities.
\newblock \emph{Remote Sensing of Environment}, 264:\penalty0 112577, 2021.

\bibitem[Lai et~al.(2017)Lai, Huang, Ahuja, and Yang]{lai2017deep}
Lai, W.-S., Huang, J.-B., Ahuja, N., and Yang, M.-H.
\newblock Deep laplacian pyramid networks for fast and accurate super-resolution.
\newblock In \emph{Proceedings of the IEEE conference on computer vision and pattern recognition}, pp.\  624--632, 2017.

\bibitem[Le et~al.(2023)Le, Zhang, Nguyen, Bolten, Lakshmi, et~al.]{le2023robustness}
Le, M.-H., Zhang, R., Nguyen, B.~Q., Bolten, J.~D., Lakshmi, V., et~al.
\newblock Robustness of gridded precipitation products for vietnam basins using the comprehensive assessment framework of rainfall.
\newblock \emph{Atmospheric Research}, 293:\penalty0 106923, 2023.

\bibitem[Li et~al.(2022)Li, Hong, Gao, Yao, Zheng, Zhang, and Chanussot]{li2022deep}
Li, J., Hong, D., Gao, L., Yao, J., Zheng, K., Zhang, B., and Chanussot, J.
\newblock Deep learning in multimodal remote sensing data fusion: A comprehensive review.
\newblock \emph{International Journal of Applied Earth Observation and Geoinformation}, 112:\penalty0 102926, 2022.

\bibitem[Liang et~al.(2021)Liang, Cao, Sun, Zhang, Van~Gool, and Timofte]{liang2021swinir}
Liang, J., Cao, J., Sun, G., Zhang, K., Van~Gool, L., and Timofte, R.
\newblock Swinir: Image restoration using swin transformer.
\newblock In \emph{Proceedings of the IEEE/CVF international conference on computer vision}, pp.\  1833--1844, 2021.

\bibitem[Liang(2005)]{liang2005quantitative}
Liang, S.
\newblock \emph{Quantitative remote sensing of land surfaces}.
\newblock John Wiley \& Sons, 2005.

\bibitem[Lim et~al.(2017)Lim, Son, Kim, Nah, and Mu~Lee]{lim2017enhanced}
Lim, B., Son, S., Kim, H., Nah, S., and Mu~Lee, K.
\newblock Enhanced deep residual networks for single image super-resolution.
\newblock In \emph{Proceedings of the IEEE conference on computer vision and pattern recognition workshops}, pp.\  136--144, 2017.

\bibitem[Liu et~al.(2020)Liu, Ding, Chen, and Liu]{liu2020similarity}
Liu, Y., Ding, L., Chen, C., and Liu, Y.
\newblock Similarity-based unsupervised deep transfer learning for remote sensing image retrieval.
\newblock \emph{IEEE Transactions on Geoscience and Remote Sensing}, 58\penalty0 (11):\penalty0 7872--7889, 2020.

\bibitem[Ma et~al.(2021)Ma, Zeng, Zhang, Fu, Zheng, Wigneron, Chen, and Niyogi]{ma2021evaluation}
Ma, H., Zeng, J., Zhang, X., Fu, P., Zheng, D., Wigneron, J.-P., Chen, N., and Niyogi, D.
\newblock Evaluation of six satellite-and model-based surface soil temperature datasets using global ground-based observations.
\newblock \emph{Remote Sensing of Environment}, 264:\penalty0 112605, 2021.

\bibitem[Ma et~al.(2019)Ma, Liu, Zhang, Ye, Yin, and Johnson]{ma2019deep}
Ma, L., Liu, Y., Zhang, X., Ye, Y., Yin, G., and Johnson, B.~A.
\newblock Deep learning in remote sensing applications: A meta-analysis and review.
\newblock \emph{ISPRS journal of photogrammetry and remote sensing}, 152:\penalty0 166--177, 2019.

\bibitem[Mahrad et~al.(2020)Mahrad, Newton, Icely, Kacimi, Abalansa, and Snoussi]{mahrad2020contribution}
Mahrad, B.~E., Newton, A., Icely, J.~D., Kacimi, I., Abalansa, S., and Snoussi, M.
\newblock Contribution of remote sensing technologies to a holistic coastal and marine environmental management framework: a review.
\newblock \emph{Remote Sensing}, 12\penalty0 (14):\penalty0 2313, 2020.

\bibitem[Messina \& Modica(2022)Messina and Modica]{messina2022twenty}
Messina, G. and Modica, G.
\newblock Twenty years of remote sensing applications targeting landscape analysis and environmental issues in olive growing: A review.
\newblock \emph{Remote Sensing}, 14\penalty0 (21):\penalty0 5430, 2022.

\bibitem[Moraux et~al.(2019)Moraux, Dewitte, Cornelis, and Munteanu]{moraux2019deep}
Moraux, A., Dewitte, S., Cornelis, B., and Munteanu, A.
\newblock Deep learning for precipitation estimation from satellite and rain gauges measurements.
\newblock \emph{Remote Sensing}, 11\penalty0 (21):\penalty0 2463, 2019.

\bibitem[Nasir et~al.(2021)Nasir, Bibi, Shah, Khan, Sharif, Iqbal, Nam, and Kadry]{nasir2021deep}
Nasir, I.~M., Bibi, A., Shah, J.~H., Khan, M.~A., Sharif, M., Iqbal, K., Nam, Y., and Kadry, S.
\newblock Deep learning-based classification of fruit diseases: An application for precision agriculture.
\newblock \emph{Comput. Mater. Contin}, 66\penalty0 (2):\penalty0 1949--1962, 2021.

\bibitem[Parker(2010)]{parker2010algorithms}
Parker, J.~R.
\newblock \emph{Algorithms for image processing and computer vision}.
\newblock John Wiley \& Sons, 2010.

\bibitem[Prodhan et~al.(2022)Prodhan, Zhang, Hasan, Sharma, and Mohana]{prodhan2022review}
Prodhan, F.~A., Zhang, J., Hasan, S.~S., Sharma, T. P.~P., and Mohana, H.~P.
\newblock A review of machine learning methods for drought hazard monitoring and forecasting: Current research trends, challenges, and future research directions.
\newblock \emph{Environmental modelling \& software}, 149:\penalty0 105327, 2022.

\bibitem[Pu \& Gong(2000)Pu and Gong]{pu2000hyperspectral}
Pu, R.-l. and Gong, P.
\newblock Hyperspectral remote sensing and its applications.
\newblock \emph{Higher Education, Beijing}, 8, 2000.

\bibitem[Ronneberger et~al.(2015)Ronneberger, Fischer, and Brox]{ronneberger2015u}
Ronneberger, O., Fischer, P., and Brox, T.
\newblock U-net: Convolutional networks for biomedical image segmentation.
\newblock In \emph{Medical image computing and computer-assisted intervention--MICCAI 2015: 18th international conference, Munich, Germany, October 5-9, 2015, proceedings, part III 18}, pp.\  234--241. Springer, 2015.

\bibitem[Sainath et~al.(2015)Sainath, Vinyals, Senior, and Sak]{sainath2015convolutional}
Sainath, T.~N., Vinyals, O., Senior, A., and Sak, H.
\newblock Convolutional, long short-term memory, fully connected deep neural networks.
\newblock In \emph{2015 IEEE international conference on acoustics, speech and signal processing (ICASSP)}, pp.\  4580--4584. Ieee, 2015.

\bibitem[Salcedo-Sanz et~al.(2020)Salcedo-Sanz, Ghamisi, Piles, Werner, Cuadra, Moreno-Mart{\'\i}nez, Izquierdo-Verdiguier, Mu{\~n}oz-Mar{\'\i}, Mosavi, and Camps-Valls]{salcedo2020machine}
Salcedo-Sanz, S., Ghamisi, P., Piles, M., Werner, M., Cuadra, L., Moreno-Mart{\'\i}nez, A., Izquierdo-Verdiguier, E., Mu{\~n}oz-Mar{\'\i}, J., Mosavi, A., and Camps-Valls, G.
\newblock Machine learning information fusion in earth observation: A comprehensive review of methods, applications and data sources.
\newblock \emph{Information Fusion}, 63:\penalty0 256--272, 2020.

\bibitem[Schneider et~al.(2010)Schneider, Friedl, and Potere]{schneider2010mapping}
Schneider, A., Friedl, M.~A., and Potere, D.
\newblock Mapping global urban areas using modis 500-m data: New methods and datasets based on ‘urban ecoregions’.
\newblock \emph{Remote sensing of environment}, 114\penalty0 (8):\penalty0 1733--1746, 2010.

\bibitem[Segal(2004)]{segal2004machine}
Segal, M.~R.
\newblock Machine learning benchmarks and random forest regression.
\newblock 2004.

\bibitem[Shi et~al.(2016)Shi, Caballero, Husz{\'a}r, Totz, Aitken, Bishop, Rueckert, and Wang]{shi2016real}
Shi, W., Caballero, J., Husz{\'a}r, F., Totz, J., Aitken, A.~P., Bishop, R., Rueckert, D., and Wang, Z.
\newblock Real-time single image and video super-resolution using an efficient sub-pixel convolutional neural network.
\newblock In \emph{Proceedings of the IEEE conference on computer vision and pattern recognition}, pp.\  1874--1883, 2016.

\bibitem[Singh et~al.(2020)Singh, Pandey, Petropoulos, Pavlides, Srivastava, Koutsias, Deng, and Bao]{singh2020hyperspectral}
Singh, P., Pandey, P.~C., Petropoulos, G.~P., Pavlides, A., Srivastava, P.~K., Koutsias, N., Deng, K. A.~K., and Bao, Y.
\newblock Hyperspectral remote sensing in precision agriculture: Present status, challenges, and future trends.
\newblock In \emph{Hyperspectral remote sensing}, pp.\  121--146. Elsevier, 2020.

\bibitem[Spadoni et~al.(2020)Spadoni, Cavalli, Congedo, and Munaf{\`o}]{spadoni2020analysis}
Spadoni, G.~L., Cavalli, A., Congedo, L., and Munaf{\`o}, M.
\newblock Analysis of normalized difference vegetation index (ndvi) multi-temporal series for the production of forest cartography.
\newblock \emph{Remote Sensing Applications: Society and Environment}, 20:\penalty0 100419, 2020.

\bibitem[Tmu{\v{s}}i{\'c} et~al.(2020)Tmu{\v{s}}i{\'c}, Manfreda, Aasen, James, Gon{\c{c}}alves, Ben-Dor, Brook, Polinova, Arranz, M{\'e}sz{\'a}ros, et~al.]{tmuvsic2020current}
Tmu{\v{s}}i{\'c}, G., Manfreda, S., Aasen, H., James, M.~R., Gon{\c{c}}alves, G., Ben-Dor, E., Brook, A., Polinova, M., Arranz, J.~J., M{\'e}sz{\'a}ros, J., et~al.
\newblock Current practices in uas-based environmental monitoring.
\newblock \emph{Remote Sensing}, 12\penalty0 (6):\penalty0 1001, 2020.

\bibitem[Tsakiris \& Loucks(2023)Tsakiris and Loucks]{tsakiris2023adaptive}
Tsakiris, G. and Loucks, D.
\newblock Adaptive water resources management under climate change: an introduction.
\newblock \emph{Water Resources Management}, 37\penalty0 (6):\penalty0 2221--2233, 2023.

\bibitem[Tsatsaris et~al.(2021)Tsatsaris, Kalogeropoulos, Stathopoulos, Louka, Tsanakas, Tsesmelis, Krassanakis, Petropoulos, Pappas, and Chalkias]{tsatsaris2021geoinformation}
Tsatsaris, A., Kalogeropoulos, K., Stathopoulos, N., Louka, P., Tsanakas, K., Tsesmelis, D.~E., Krassanakis, V., Petropoulos, G.~P., Pappas, V., and Chalkias, C.
\newblock Geoinformation technologies in support of environmental hazards monitoring under climate change: An extensive review.
\newblock \emph{ISPRS International Journal of Geo-Information}, 10\penalty0 (2):\penalty0 94, 2021.

\bibitem[Wang et~al.(2018)Wang, Yu, Wu, Gu, Liu, Dong, Qiao, and Change~Loy]{wang2018esrgan}
Wang, X., Yu, K., Wu, S., Gu, J., Liu, Y., Dong, C., Qiao, Y., and Change~Loy, C.
\newblock Esrgan: Enhanced super-resolution generative adversarial networks.
\newblock In \emph{Proceedings of the European conference on computer vision (ECCV) workshops}, pp.\  0--0, 2018.

\bibitem[Wang et~al.(2023)Wang, Meili, and Fatichi]{wang2023evidence}
Wang, Y., Meili, N., and Fatichi, S.
\newblock Evidence and controls of the acceleration of the hydrological cycle over land.
\newblock \emph{Water Resources Research}, 59\penalty0 (8):\penalty0 e2022WR033970, 2023.

\bibitem[Wang et~al.(2004)Wang, Bovik, Sheikh, and Simoncelli]{wang2004image}
Wang, Z., Bovik, A.~C., Sheikh, H.~R., and Simoncelli, E.~P.
\newblock Image quality assessment: from error visibility to structural similarity.
\newblock \emph{IEEE transactions on image processing}, 13\penalty0 (4):\penalty0 600--612, 2004.

\bibitem[Willmott \& Matsuura(2005)Willmott and Matsuura]{willmott2005advantages}
Willmott, C.~J. and Matsuura, K.
\newblock Advantages of the mean absolute error (mae) over the root mean square error (rmse) in assessing average model performance.
\newblock \emph{Climate research}, 30\penalty0 (1):\penalty0 79--82, 2005.

\bibitem[Willmott et~al.(2012)Willmott, Robeson, and Matsuura]{willmott2012refined}
Willmott, C.~J., Robeson, S.~M., and Matsuura, K.
\newblock A refined index of model performance.
\newblock \emph{International Journal of climatology}, 32\penalty0 (13):\penalty0 2088--2094, 2012.

\bibitem[Wu et~al.(2020)Wu, Yang, Liu, and Wang]{wu2020spatiotemporal}
Wu, H., Yang, Q., Liu, J., and Wang, G.
\newblock A spatiotemporal deep fusion model for merging satellite and gauge precipitation in china.
\newblock \emph{Journal of Hydrology}, 584:\penalty0 124664, 2020.

\bibitem[Wu et~al.(2019)Wu, Xiao, Wen, You, and Hueni]{wu2019advances}
Wu, X., Xiao, Q., Wen, J., You, D., and Hueni, A.
\newblock Advances in quantitative remote sensing product validation: Overview and current status.
\newblock \emph{Earth-Science Reviews}, 196:\penalty0 102875, 2019.

\bibitem[Xia et~al.(2023)Xia, Zhang, Wang, Wang, Wu, Tian, Yang, and Van~Gool]{xia2023diffir}
Xia, B., Zhang, Y., Wang, S., Wang, Y., Wu, X., Tian, Y., Yang, W., and Van~Gool, L.
\newblock Diffir: Efficient diffusion model for image restoration.
\newblock In \emph{Proceedings of the IEEE/CVF International Conference on Computer Vision}, pp.\  13095--13105, 2023.

\bibitem[Xu et~al.(2022)Xu, Huang, Si, Li, Xu, Zhang, Song, and Gao]{xu2022onboard}
Xu, H., Huang, W., Si, X., Li, X., Xu, W., Zhang, L., Song, Q., and Gao, H.
\newblock Onboard spectral calibration and validation of the satellite calibration spectrometer on hy-1c.
\newblock \emph{Optics Express}, 30\penalty0 (15):\penalty0 27645--27661, 2022.

\bibitem[Yuan et~al.(2020)Yuan, Shen, Li, Li, Li, Jiang, Xu, Tan, Yang, Wang, et~al.]{yuan2020deep}
Yuan, Q., Shen, H., Li, T., Li, Z., Li, S., Jiang, Y., Xu, H., Tan, W., Yang, Q., Wang, J., et~al.
\newblock Deep learning in environmental remote sensing: Achievements and challenges.
\newblock \emph{Remote sensing of Environment}, 241:\penalty0 111716, 2020.

\bibitem[Zhang et~al.(2021)Zhang, Liang, Van~Gool, and Timofte]{zhang2021designing}
Zhang, K., Liang, J., Van~Gool, L., and Timofte, R.
\newblock Designing a practical degradation model for deep blind image super-resolution.
\newblock In \emph{Proceedings of the IEEE/CVF International Conference on Computer Vision}, pp.\  4791--4800, 2021.

\bibitem[Zhao et~al.(2022)Zhao, Yu, Du, Peng, Hao, Zhang, and Gong]{zhao2022overview}
Zhao, Q., Yu, L., Du, Z., Peng, D., Hao, P., Zhang, Y., and Gong, P.
\newblock An overview of the applications of earth observation satellite data: impacts and future trends.
\newblock \emph{Remote Sensing}, 14\penalty0 (8):\penalty0 1863, 2022.

\bibitem[Zhou \& Liu(2024)Zhou and Liu]{zhou2024enhancing}
Zhou, L. and Liu, L.
\newblock Enhancing dynamic flood risk assessment and zoning using a coupled hydrological-hydrodynamic model and spatiotemporal information weighting method.
\newblock \emph{Journal of Environmental Management}, 366:\penalty0 121831, 2024.

\end{thebibliography}
\bibliographystyle{icml2025}





\end{document}